\documentclass[10pt, a4paper]{article}

\usepackage{lrec-coling2024} 

\usepackage{multirow}
\usepackage{natbib}  
\usepackage{svg}
\usepackage{multibib}
\usepackage{array}
\makeatletter
\def\@mb@citenamelist{cite,citep,citet,citealp,citealt,citepalias,citetalias}
\makeatother
\newcites{languageresource}{~}

\usepackage{graphicx}
\usepackage{tabularx}
\usepackage{soul}
\usepackage{xcolor}

\usepackage{caption}
\usepackage{subcaption}
\usepackage{titlesec}
\titleformat{\section}{\normalfont\large\bfseries\center}{\thesection.}{1em}{}
\titleformat{\subsection}{\normalfont\SmallTitleFont\bfseries\raggedright}{\thesubsection.}{1em}{}
\titleformat{\subsubsection}{\normalfont\normalsize\bfseries\raggedright}{\thesubsubsection.}{1em}{}
\renewcommand\thesection{\arabic{section}}
\renewcommand\thesubsection{\thesection.\arabic{subsection}}
\renewcommand\thesubsubsection{\thesubsection.\arabic{subsubsection}}

\usepackage{xcolor}
\usepackage{hyperref}
 \definecolor{darkblue}{rgb}{0, 0, 0.5}
  \hypersetup{colorlinks=true, citecolor=darkblue, linkcolor=darkblue, urlcolor=darkblue}

\usepackage{xstring}

\usepackage{color}

\usepackage{placeins}

\title{Pointing out the Shortcomings of Relation Extraction Models with Semantically Motivated Adversarials}

\name{Gennaro Nolano, $^1$Moritz Blum, $^{1,2}$Basil Ell, $^1$Philipp Cimiano} 
\address{$^1$Bielefeld University, $^2$University of Oslo \\
        Germany, Norway \\
         \{mblum, bell, cimiano\}@techfak.uni-bielefeld.de, nolanogenn@gmail.com\\}

\abstract{In recent years, large language models have achieved state-of-the-art performance across various NLP tasks. However, investigations have shown that these models tend to rely on shortcut features, leading to inaccurate predictions 
and causing the models to be unreliable at generalization to out-of-distribution~(OOD) samples. For instance, in the context of relation extraction~(RE), we would expect a model to identify the same relation independently of the entities involved in it. For example, consider the sentence \textit{"Leonardo da Vinci painted the Mona Lisa"} expressing the \texttt{created(Leonardo\_da\_Vinci, Mona\_Lisa)} relation. If we substiute \textit{"Leonardo da Vinci"} with \textit{"Barack Obama"}, then the sentence still expresses the \texttt{created} relation
. A robust model is supposed to detect the same relation in both cases.\newline
In this work, we describe several semantically-motivated strategies to generate adversarial examples by replacing entity mentions and investigate how state-of-the-art RE models perform under pressure. 
Our analyses show that the performance of these models significantly deteriorates on the modified datasets~(avg. of -48.5\% in $F1$), which indicates that these models rely to a great extent on shortcuts, such as surface forms~(or patterns therein) of entities, without making full use of the information present in the sentences.
\\ \newline 
}

\begin{document}

\maketitleabstract

\section{Introduction}\label{sec:introduction}Recent large language models have achieved state-of-the-art performance on many NLP tasks, such as Relation Extraction~(RE), and are generally considered as the go-to methodology to tackle this sort of task~\cite{ji:22:survey}. These models exploit contextual information extracted from natural language sentences in the form of embeddings to label relations between entity mentions~\cite{baldini-soares-etal-2019-matching,wu:19:rbert}.\newline
Despite achieving generally good results, closer investigations show that these modes' inner workings have yet to be fully understood, and that they are less robust than expected.For instance, the work by \citet{rosenman-etal-2020-exposing} highlights that SOTA models for RE seem to exploit simple heuristics. Furthermore, these models show a lack of robustness when put under pressure \cite{tenney-etal-2020-language}, such as under domain shift~\cite{blitzer:08} and in adversarial settings~\cite{papernot2016crafting, jia-liang-2017-adversarial, ebrahimi-etal-2018-hotflip, belinkov2017synthetic}. \newline
In this work, we investigate the behavior of models fine-tuned for RE in an adversarial setting, showing that they act in unexpected ways when encountering unforeseen situations. In particular, we test the models using several adversarial datasets generated by substituting the entities mentioned in a sentence realizing a specific relation, and by following specific strategies. In doing so, the contextual information between and around the entities, and the relation expressed by the sentence, is left untouched.\newline
For instance, considering the sentence expressing a relation between the entities \textit{Leonardo Da Vinci}~($e_{SUBJ}$) and the \textit{Mona Lisa}~($e_{OBJ}$):
\begin{center}
\noindent 
"\textit{\underline{Leonardo Da Vinci} painted the \underline{Mona Lisa}.}"
\end{center}
\noindent Replacing the entity \textit{Leonardo Da Vinci} with another entity, such as \textit{Michelangelo}, \textit{Barack Obama} or \textit{Stratolaunch}, will inevitably change the semantics of the sentence, but a robust RE system should still recognize the original relation to be expressed given that, beyond the entities, the actual context is unchanged.\newline
We hypothesize that, in case a model is not able to predict the correct relation type given such adversarial examples, it is because the model itself has learned entity- and relation-specific heuristics at training time, rather than actually learning how relations are expressed in natural language.\newline
More specifically, given a text example expressing a relation $r$ between entities $e_{SUBJ}$ and $e_{OBJ}$, we replace $e_{SUBJ}$, $e_{OBJ}$, or both, with other entities, according to specific substitution operations.\newline
Our experiments with different types of substitution operations show that the models are significantly misled by such adversarials, reducing performance by an average of $48.5\%$ in \textit{F1}, depending on the model involved and the substitution operations applied. 

\section{Related Work}\label{sec:rw}Shortcut learning in NLP refers to the process where a model relies on data biases to make predictions. More precisely, the model learns a co-occurrence of certain features and labels, without achieving true language understanding. These features are called shortcut features. \citet{du-etal-2021-towards} demonstrate that NLP models have a strong preference for features located at the head of the long-tailed local mutual information~(LMU)~\cite{schuster-etal-2019-towards} distribution of words and labels. If a model learns such shortcut features, it might suffer from low generalization abilities and low adversarial robustness on data that do not share the same shortcut features as the training data. 

To explore an NLP model's inclination towards shortcut learning, one approach is to examine how the model behaves when confronted with adversarials~\cite{morris-etal-2020-textattack}. Other approaches analyze syntactic heuristics~\cite{mccoy-etal-2019-right} or do randomization ablation studies~\cite{sinha2021masked}.

The core idea behind the use of adversarials to test ML models has its roots in the field of computer vision, where the incorporation of small perturbations into input images has been proposed as a way to create difficult-to-solve datasets \cite{szegedy:14, goodfellow:15}. Similar frameworks have been implemented for NLP tasks as well by employing different kinds of permutations \cite{Goyal2022ASO}.

Generally speaking, adversarial attacks can be divided into two categories according to \citet{gainski-balazy-2023-step}: white-box methods, which make use of gradients to generate perturbations \cite{ebrahimi-etal-2018-hotflip, cheng-etal-2019-robust, Xu2020TextTrickerLA,yuan2023bridge}, and black-box methods, which define heuristics to implement input data permutations, e.g. character or word replacements \cite{yoo-etal-2020-searching,morris-etal-2020-textattack}. The present work implements a black-box methodology with heuristics based on the semantics of relations and entity types. For a list of adversarial attack methodologies and techniques for NLP, see \cite{morris-etal-2020-textattack}.

Adversarials have been widely used to highlight specific points of weakness in NLP models, e.\thinspace g., \citet{li:16} use feature erasure to explain neural model decisions over several tasks, such as POS tagging and word frequency prediction. Similarly, \citet{ribeiro-etal-2018-semantically} use adversarials to investigate points of failure in machine comprehension, visual QA, and sentiment analysis.\par
Similarly, \citet{hosseini2017deceiving} have shown the effect of symbol addition and typo insertion on the task of toxic language detection, while \citet{kaushik-lipton-2018-much} show that reading comprehension datasets are not heavily affected by the removal of questions. Finally, \citet{belinkov2017synthetic} applied such permutations to investigate the robustness of machine translation.\par
Recently, \citet{formento-etal-2023-using} proved the effectiveness of character insertion as a form of adversarial attack against several tasks, such as text classification, entailment, and question answering.

While these models prove the effects of using adversarials to investigate NLP models, they generally exploit simple surface patterns to generate adversarials \cite{wallace-etal-2019-universal}. A more complex framework in this regard is proposed by \citet{li:20:bert}, who make use of entity-altering permutations to investigate the robustness of \textit{BERT}-based models for RE. However, they only investigate the substitution of entities with entities of the same semantic type, and entity masking. Moreover, they limit their evaluation to BERT-based models.\par
Beyond existing works and in particular the work of \citet{li:20:bert}, we consider the impact of adversarials in RE by examining the effect of more complex types of entity substitutions, while also taking into account the performance of various state-of-the-art models for RE.

\section{Methodology}\label{sec:methodology}We construct $12$ adversarial datasets by changing the semantics of sentences while preserving the underlying relation between entities expressed in the sentence.\newline
We argue that, while information about the entities involved in the relation is relevant to determine whether that relation can exist between the two entities, a robust model should not rely only on information about the entities, such as their semantic types or the gender of a mentioned person, but should make use of the intra-sentence context in which the entities are mentioned.\newline
In general, human readers can identify the relation between two entities as expressed in a natural language sentence by following simple lexico-syntactic patterns \cite{hearst-1992-automatic}, and the same should be expected from a RE model.\newline
We obtain adversarial examples from a sentence that expresses a specific relation (e.\thinspace g., \texttt{created} in the example about the \textit{Mona Lisa} shown in Sec.~\ref{sec:introduction}) between a subject entity and an object entity by replacing the mention of the subject or object entity (or both) with the surface form of another entity. Thereby, from a sentence that expresses a relation that is factually true (e.\thinspace g., \textit{"Leonardo Da Vinci painted the Mona Lisa."}) we may obtain a sentence that is factually incorrect (e.\thinspace g., \textit{"Leonardo Da Vinci painted the Scream."}). However, the relation expressed (\texttt{created}) stays the same, albeit between different entities.\newline
We introduce four substitution strategies which we apply to create 12 adversarial datasets to test how models fare under different semantically-motivated permutations.\newline
We define our corpus $D$ as a set of quadruples of the form $(d,r,e_{SUBJ},e_{OBJ})$ where $d$ is a sentence expressing the relation $r$ between the subject entity $e_{SUBJ}$ and the object entity $e_{OBJ}$. By $\lambda_{SUBJ}$ and $\lambda_{OBJ}$ we denote the mention of the subject and object entitity in the sentence $d$, respectively, and by $t_{SUBJ}$ and $t_{OBJ}$ we denote the type of the subject and object entitity, respectively.\newline
Given a quadruple $(d,r,e_{SUBJ},e_{OBJ})$, each substitution strategy creates a new quadruple $(d',r,e_{SUBJ}',e_{OBJ}')$ where either the original subject entity is replaced by another entity, the original object entity is replaced by another entity, or both are replaced, and where a new sentence $d'$ is obtained by replacing the original surface form of the previously mentioned entity(ies) with the surface form(s) of the new entity(ies).\newline
The substitution strategies are as follows:
\begin{itemize}
\setlength\itemsep{0em}
    \item{\textbf{same-role substitution}}

    $e_{SUBJ}'$ is randomly chosen from the set
    $\lbrace x ~|~ \exists (d'', r, x, e_{OBJ}'') \in D : x \neq e_{SUBJ} \rbrace$ and $e_{OBJ} = e_{OBJ}'$. $e_{OBJ}'$ can be selected analogously.

    Thus, an entity is replaced with another entity that occurs in the same role (as subject or object) in another sentence expressing the same relation. Either the subject or the object or both are replaced.
    
    \item{\textbf{same-type substitution}}

    $e_{SUBJ}'$ is randomly chosen from the set
    $\lbrace x ~|~ \exists (d'', r', x, e_{OBJ}'') \in D : r \neq r' \land x \neq e_{SUBJ} \land t_{SUBJ} = t_{SUBJ}'\rbrace$. $e_{OBJ}'$ can be selected analogously.

    Thus, an entity is replaced with another entity of the same type that occurs in the same role (as subject or object) in another sentence expressing another relation. Either the subject or the object or both are replaced.

    %
    \item{\textbf{different-type substitution}}
    $e_{SUBJ}'$ is randomly chosen from the set
    $\lbrace x ~|~ \exists (d'', r', x, e_{OBJ}'') \in D : r \neq r' \land x \neq e_{SUBJ} \land t_{SUBJ} \neq t_{SUBJ}'\rbrace$. 
    $e_{OBJ}'$ can be selected analogously.

    Thus, an entity is replaced with another entity of a different type that never occurs in the same role (as subject or object) in another sentence expressing the same relation. Either the subject or the object or both are replaced.
    
   \item{\textbf{masking}}

    $e_{SUBJ}'$ is replaced with the \texttt{[MASK]} token. 
    We set $t_{SUBJ}'$ to \texttt{NONE}. 
    $e_{OBJ}'$ can be selected analogously.
    Either the subject or the object or both are replaced.
   
   \end{itemize}
We apply each of the four strategies to replace either the subject, the object or both entities to each sentence of the testing set, thus obtaining $12$ adversarial datasets from the original data. 
An original sentence and one adversarial example for each substitution strategy is shown in Table~\ref{tab:adv_examples}.
\begin{table}[!ht]
\scriptsize
    \caption{Examples of the adversarial strategies}
    \label{tab:adv_examples}
    \centering
    \begin{tabular}{llll}
    & \multicolumn{3}{c}{\textbf{Original Sentence}}\\
    \hline
    & Leonardo da Vinci & painted the & Mona Lisa \\
    \hline
    \hline
    & \multicolumn{3}{c}{\textbf{Same-role}}\\
    \hline
    \textbf{subj mod}. & Michelangelo & painted the & Mona Lisa \\
    \textbf{obj mod}. & Leonardo da Vinci & painted the & Scream \\
    \textbf{subj+obj mod}. & Michelangelo & painted the & Scream \\
    \hline
    \hline
    & \multicolumn{3}{c}{\textbf{Same-type}}\\
    \hline
    \textbf{subj mod}. & Barack Obama & painted the & Mona Lisa \\
    \textbf{obj mod}. & Leonardo da Vinci & painted the & Baloon Girl \\
    \textbf{subj+obj mod}. & Barack Obama & painted the & Baloon Girl \\
    \hline
    \hline
    & \multicolumn{3}{c}{\textbf{Diff.-type}}\\
    \hline
    \textbf{subj mod}. & Stratolaunch & painted the & Mona Lisa \\
    \textbf{obj mod}. & Leonardo da Vinci & painted the & Berlin Wall \\
    \textbf{subj+obj mod}. & Stratolaunch & painted the & Berlin Wall \\
    \hline
    \hline
    & \multicolumn{3}{c}{\textbf{Masking}}\\
    \hline
    \textbf{subj mod}. & [MASK] & painted the & Mona Lisa \\
    \textbf{obj mod}. & Leonardo da Vinci & painted the & [MASK] \\
    \textbf{subj+obj mod}. & [MASK] & painted the & [MASK] \\
    \hline
    \hline
    \end{tabular}

\end{table}

\section{Experiments}\label{sec:experiment}\subsection{Data}
In order to build adversarial datasets and to test the models, the starting point is the human-labelled TACRED dataset~\cite{zhang-etal-2017-position}. The dataset contains 68,124 samples for training, 22,631 samples for evaluation, and 15,509 samples for testing, expressing 42 relation types. The dataset only contains English sentences.\newline
In each sample, both entities are labelled with their semantic type,\footnote{From the Stanford NER system, see \url{https://stanfordnlp.github.io/CoreNLP/ner.html}} and linked, when possible, to their corresponding entry in Wikidata.\footnote{Wikidata version accessed \textit{Wikidata 2022-01-03}}\newline
We use the training split where needed to finetune the models and use the test split to generate the adversarials examples used to test the models.\footnote{The script to create the proposed adversarial datasets from TACRED will be published after the anonymity period is over.}\newline
Some sentences in TACRED contain arguments of relations that cannot be linked to Wikidata. This is, for example, the case when the mention is a personal pronoun. By this process, we focus on sentences where entities are mentioned explicitly and ignore sentences where the entities are only indirectly mentioned / referred to. We discard those sentences in the test split in which at least on argument of the relation in not linked to Wikidata.\newline 
This process removes 9,232 sentences from the test split, thus leaving us with 6,277 sentences that were then used to generate adversarial examples.\newline
One of the most peculiar aspects of TACRED is that the relation classes are not equally distributed. In fact, the training set is, by design, heavily skewed towards the \texttt{no\_relation} label,\footnote{See the official dataset statistics at \url{https://nlp.stanford.edu/projects/tacred/\#stats}} which heavily affects models as they seem to default to this class when facing unexpected pairs of entities, as described in Section~\ref{subsec:rel}.
\subsection{Models}
The adversarial examples are tested using different SOTA models on TACRED. In particular, the investigated models are: \textbf{LUKE} \cite{yamada-etal-2020-luke}, \textbf{SpanBERT} \cite{joshi2020spanbert}, \textbf{UniST} \cite{huang-etal-2022-unified}, \textbf{SuRE} \cite{lu2022summarization}, \textbf{TYP-Marker} \cite{zhou-chen-2022-improved}, and \textbf{NLI} \cite{sainz-rigau-2021-ask2transformers}. For LUKE and SpanBERT, the checkpoints are loaded from the HuggingfaceHub,\footnote{\url{https://huggingface.co/studio-ousia/luke-large}}\textsuperscript{,}\footnote{\url{https://huggingface.co/mrm8488/spanbert-base-finetuned-tacred}} while the checkpoints for NLI are downloaded following the instructions found in the repository.\footnote{\url{http://github.com/osainz59/Ask2Transformers}} Since no weights are publicly available for SuRE and UniST, we finetune these models using the TACRED training set by following the procedures described in the official repositories\footnote{\url{https://github.com/luka-group/unist}}\textsuperscript{,}\footnote{\url{https://github.com/luka-group/sure}} before testing them in the adversarial setting.\newline
For the NLI model, we test both the original model (\textbf{NLI\_w}), which implements heuristics-based constraints on the possible entity types that can appear as subject/object for a given relation, and a version of the model where these constraints are not implemented (\textbf{NLI\_w/o}).

\section{Results}\label{sec:results}\subsection{General Results}\label{subsec:general_results}
We evaluate the performance of the models under pressure using the \textit{F1} score computed with the official TACRED evaluation script.\footnote{\url{https://github.com/yuhaozhang/tacred-relation/tree/master}} The \textit{F1} values for every model and each of our adversarial strategies/datasets are shown in Table~\ref{tab:resultsf1}.

\begingroup

\setlength{\tabcolsep}{4.3pt} 

\begin{table*}[!ht]
{
    \centering
    \scriptsize
    \caption{Fine-grained \textit{F1} scores on \textit{Tacred}. \textbf{std.} the evaluation on the standard test set, \textbf{adv.} the average evaluation over all the adversarial strategies, \textbf{diff.} the percentage of loss from standard to adversarial evaluation. The next columns contain \textit{F1} score for each strategy.}
    \begin{tabular}
      {
    |r||
    >{\raggedleft\arraybackslash}p{0.6cm}|
    >{\raggedleft\arraybackslash}p{0.6cm}|
    >{\raggedleft\arraybackslash}p{0.8cm}||
    >{\raggedleft\arraybackslash}p{0.6cm}|
    >{\raggedleft\arraybackslash}p{0.6cm}|
    >{\raggedleft\arraybackslash}p{0.6cm}||
    >{\raggedleft\arraybackslash}p{0.6cm}|
    >{\raggedleft\arraybackslash}p{0.6cm}|
    >{\raggedleft\arraybackslash}p{0.6cm}||
    >{\raggedleft\arraybackslash}p{0.6cm}|
    >{\raggedleft\arraybackslash}p{0.6cm}|
    >{\raggedleft\arraybackslash}p{0.6cm}||
    >{\raggedleft\arraybackslash}p{0.6cm}|
    >{\raggedleft\arraybackslash}p{0.6cm}|
    >{\raggedleft\arraybackslash}p{0.6cm}|
    }
    \hline
        \multirow{2}{*}{\textbf{Model}}
    &   \multirow{2}{*}{\textbf{std.}}
    &   \multirow{2}{*}{\textbf{adv.}} 
    &   \multirow{2}{*}{\textbf{diff.}} 
    &   \multicolumn{3}{l||}{\textbf{same-role sub.}}
    &   \multicolumn{3}{l||}{\textbf{same-type sub.}}
    &   \multicolumn{3}{l||}{\textbf{diff.-type sub.}}
    &   \multicolumn{3}{l|}{\textbf{masking sub.}} \\
    & & & 
    & subj 
    & obj 
    & subj + obj 
    & subj 
    & obj 
    & subj + obj 
    & subj 
    & obj 
    & subj + obj 
    & subj 
    & obj 
    & subj + obj \\
    \hline
    \hline
    LUKE            & 72.0  & 54.2   & -24.7\%   & 69.2 & 65.5  & 64.9  & 67.8  & 60.7  & 57.3  & 60.9  & 35.0  & 31.7  & 66.7  & 43.1  & 27.7 \\
    SpanBERT   & 70.8  & 26.1   & -63.0\%  & 42.4 & 41.7  & 39.8  & 39.4  & 40.3  & 35.1  & 35.0  & 32.3  & 22.9  & 30.6  & 37.4  & 23.5  \\   
    UniST           & 75.5  & 33.4   & -39.4\%   & 53.2 & 47.4  & 47.5  & 49.9  & 40.3  & 35.7  & 40.0  & 15.3  & 8.3  & 46.1  & 13.6  & 3.6  \\
    SuRE            & 74.8  & 22.4   & -59.9\%   & 37.7 & 32.5  & 28.9  & 36.5  & 29.2  & 23.5  & 29.8  & 12.5  & 8.1   & 24.8  & 5.7   & 0.1  \\
    TYP-Marker      & 72.0  & 50.6   & -29.7\%   & 68.7 & 64.5  & 63.3  & 64.0  & 57.1  & 48.4  & 52.4  & 26.4  & 15.5  & 66.0  & 44.6  & 36.0  \\
    NLI (w/)         & 68.6  & 30.4   & -55.6\%   & 63.4 & 58.4  & 54.8  & 53.6  & 52.1  & 38.2  & 24.4  & 14.9  & 4.8   & 0.0   & 0.0   & 0.0 \\
    NLI (w/o)       &42.7&27.9&-34.4\%&42.6&40.8&38.5&42.6& 39.5&36.5&37.4&24.3& 21.0&0.3&0.1&0.1 \\
    \hline
    \hline
    \textbf{avg}    &68.0&35.0&-48.5\%&53.8&50.1&48.1&50.5&45.6&39.2&40.1&22.9&16.0&33.5&20.6&14.5 \\
    \hline
    \end{tabular}
    \label{tab:resultsf1}
    }
\end{table*}

\endgroup 
The results show that the models are affected by the adversarials, with an average loss of $48.5\%$ in \textit{F1} score. As it turns out, not all adversarial strategies have the same impact on a model's performance: in particular, the \textit{same-role} and the \textit{same-type} substitution are the least impactful ones, while the \textit{masking} substitution has the strongest impact on the results, which proves that models strongly rely on entity surface forms to predict relation classes, and are led astray when such entity surface forms are absent.\newline 
It seems that models learn that they can rely on the surface forms instead of paying attention to how the relation is expressed in the context of the surface forms with which the entities are mentioned. Thus, the models perform weakly in case of unexpected pairs of entities (see Section \ref{subsec:rel}).\newline
We also observe that, on average, model performance follows a similar pattern with regards to the adversarial strategy applied. In particular, the models generally fare better when the subject entity, rather than the object entity, is substituted. The only exception is SpanBERT, where replacing either the subject entity or the object entity results in a comparable decrease in performance.\newline
We also observe that substituting both entities has the strongest negative impact for the first three types of substitutions. Among the models tested, only the UniST model exhibits a marginal performance improvement when both entities are replaced in a \textit{same-role} substitution compared to when only object is replaced. On average, the \textit{F1} score is $9.67$ points lower when objects are substituted, than when subjects are substituted, with difference as high as $17.2$ \textit{F1} score in the case of \textit{diff.-type} substitution.\newline
The most robust model, i.\thinspace e., LUKE, which reaches $72.0$ \textit{F1} in the standard evaluation, loses an average of $24.7\%$ of \textit{F1} in the adversarial evaluation. Even in cases of adversarials that are semantically closer to the original sentence, i.\thinspace e., \textit{same-role} and \textit{same-type}, this model reaches as low as $64.9$ and $57.3$ of \textit{F1}, respectively, and in case of \textit{masking} entities, it reaches as low as $27.7$ \textit{F1} score.\newline
Some models fare particularly bad in the \textit{masking} strategy. Here, NLI (w/o) reaches an \textit{F1} score between $0.0$ and $0.27$ when entities are masked. NLI (w/) always reaches $0.0$ \textit{F1} score in the \textit{masking} substitutions.
\subsection{Relations predicted}\label{subsec:rel}
The models, when put under pressure, tend to default to the \texttt{no\_relation} label, which turns out to be the most frequently predicted relation. While this is to be expected given the unbalanced nature of TACRED, the models still predict \texttt{no\_relation} more frequently in the adversarial setting, compared to those actually present in the data, as shown in Table~\ref{tab:no_rel}.
In this case, as well, object substitutions have a stronger impact than subject substitutions. In fact, across all cases except for the \textit{same-type} substitution (SpanBERT, NLI (w/)) and the \textit{masking} substitution (SpanBERT, NLI (w/o)), object substitutions lead the models to predict the \texttt{no\_relation} label with equal or greater frequency compared to subject substitutions.
\begingroup

\setlength{\tabcolsep}{2.5pt} 
\begin{table*}[!ht]
{
    \scriptsize
    \caption{Changes in predicted \textit{no\_relation} labels compared to actual ones, for every model across all adversarial datasets and testing set. The \textbf{std.} column refers to the standard test set.}
    \label{tab:no_rel}
    \begin{tabular}
    {
    |r||
    >{\raggedleft\arraybackslash}p{0.7cm}||
    >{\raggedleft\arraybackslash}p{0.9cm}|
    >{\raggedleft\arraybackslash}p{0.9cm}|
    >{\raggedleft\arraybackslash}p{0.9cm}||
    >{\raggedleft\arraybackslash}p{0.9cm}|
    >{\raggedleft\arraybackslash}p{0.9cm}|
    >{\raggedleft\arraybackslash}p{0.9cm}||
    >{\raggedleft\arraybackslash}p{0.9cm}|
    >{\raggedleft\arraybackslash}p{0.9cm}|
    >{\raggedleft\arraybackslash}p{0.9cm}||
    >{\raggedleft\arraybackslash}p{0.9cm}|
    >{\raggedleft\arraybackslash}p{0.9cm}|
    >{\raggedleft\arraybackslash}p{0.9cm}|
    }
    \hline
        \multirow{2}{*}{\textbf{Model}}
    &   \multirow{2}{*}{\textbf{std.}}
    &   \multicolumn{3}{l||}{\textbf{same-role sub.}}
    &   \multicolumn{3}{l||}{\textbf{same-type sub.}}
    &   \multicolumn{3}{l||}{\textbf{diff.-type sub.}}
    &   \multicolumn{3}{l|}{\textbf{masking sub.}} \\ 
    & 
    & subj 
    & obj 
    & subj + obj 
    & subj 
    & obj 
    & subj + obj 
    & subj 
    & obj 
    & subj + obj 
    & subj 
    & obj 
    & subj + obj \\
    \hline
    \hline
    LUKE            & -1.7\%  & +4.0\% & +6.4\%  & +8.0\%  & +2.0\%  & +6.5\%  & +8.4\%  & +6.4\%  & +17.8\%  & +18.7\%  & +2.8\%  & +10.9\%  & +24.7\% \\
    SpanBERT   & +7.0\%  & +12.4\% & +14.2\%  & +17.7\%  & +13.3\%  & +12.3\% & +17.3\%  & +15.7\%  & +18.7\%  & +25.5\%  & +19.2\%  & +8.7\%  & +24.1\%  \\   
    UniST           & +1.0\%  & +5.7\% & +8.9\%  & +11.2\%  & +6.3\%  & +9.7\%  & +23.2\%  & +28.4\%  & +28.5\%  & +35.3\%  & +39.2\%  & +39.2\%  & +39.2\%  \\
    SuRE            & -1.1\%  & +26.2\% & +28.4\%  & +30.3\%  & +27.6\%  & +28.9\%  & +31.3\%  & +29.0\%  & +34.8\%  & +36.0\%   & +31.6\%  & +37.8\%   & +38.8\%  \\
    TYP-Marker      & +1.3\%  & +6.9\% & +9.4\%  & +11.6\%  & +6.8\%  & +10.4\%  & +13.3\%  & +13.0\%  & +21.8\%  & +28.1\%  & +6.6\%  & +20.6\%  & +23.8\%  \\
    NLI (w/)         & +3.3\%  & +13.5\% & +16.0\%  & +19.6\%  & +16.6\%  & +15.7\%  & +23.2\%  & +28.4\%  & +28.5\%  & +35.3\% & +39.2\% & +39.2\%   & +39.2\%\\
     NLI (w/o)         & +3.2\%  & +11.8\% & +14.3\%  & +18.4\%  & +10.6\%  & +12.9\%  & +16.8\%  & +10.6\%  & +22.6\%  & +27.2\% & +39.1\% & +38.7\%   & +8.9\%\\
    \hline
    \end{tabular}
    }
\end{table*}
\endgroup 

\begin{figure*}
\centering
         \caption{Comparison: LUKE's predictions on the standard test set vs. the predictions on the test set following the \textit{diff.-type} object substitution strategy for selected relations.}
         \label{fig:sankey} 
        \includegraphics[scale=0.7,trim=55 50 60 50, clip]{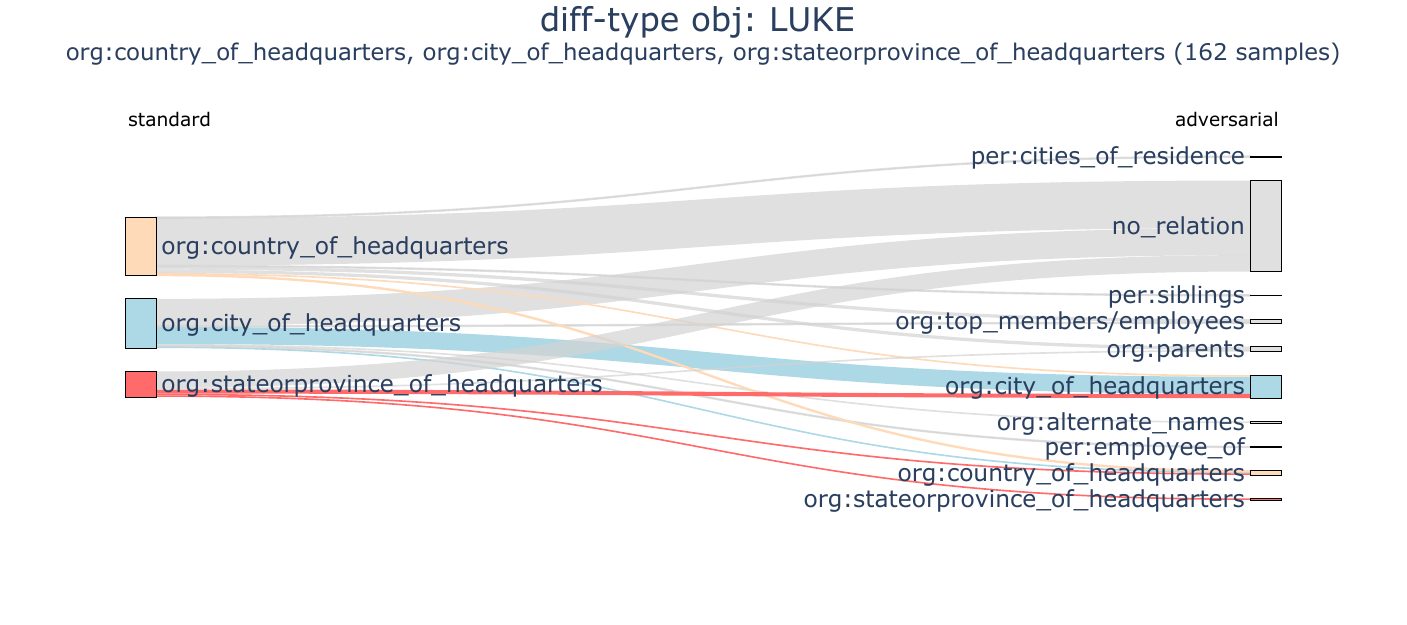}
\end{figure*}

In view of the above, we can assume that one of the main sources of inaccuracy is that models default to the most frequent label in the training set (in this case to \texttt{no\_relation} by a large margin) when put under unexpected situations.\newline
Furthermore, as depicted in Figure~\ref{fig:sankey}, a comparison is made between the predictions on the test set and those on the test set following \textit{diff.-type} object substitutions, with LUKE serving as an exemplary model. The figure shows that the model, beyond predicting the \texttt{no\_relation} label very frequently, also predicts a significant number of random relations.\newline
This behavior was consistently observed across all models examined. Additionally, despite the high similarity of the relations (with distinctions primarily in the relation scope), the model's performance is notably superior for \texttt{org:\-city\-\_of\_headquarters} relation compared to \texttt{org:country\-\_of\_headquarters} and \texttt{org:state\-orprovince\_of\_headquarters}. The cause of these phenomena remains uncertain. \newline
Nevertheless, it is important to note that they are not attributable to a bias in the number of training samples.\footnote{We generated plots for all sets of mutually confusable relations for all substitution strategies and models. These plots are generated with our referenced code and will be published after the anonymity period is over.}
\newline
It is possible to imagine the situation that a model correctly identifies that $r$ is the only relation that is expressed in a sentence $d$ and then, after an entity in $d$ was replaced with another entity, resulting in the sentence $d'$, it is no mistake if the model identifies that the only relation that is expressed in $d'$ is a relation $r'$ which is different from $r$. 
For example, in a sentence like \textit{"He lives in \underline{Italy}"} that expresses the \texttt{per:country\_of\_residence} relation, it can be appropriate that after a country is replaced by a city, then the relation that is expressed is \texttt{per:cities\_of\_residence}. 
Here, the context of the replaced entity remains the same, the same context might be equally indicative for each of the two relations. 

These situations need to be taken into account. Otherwise, one would penalize a model for predicting the actual relation being expressed in the context of the sentence.
We carry out a case study to investigate whether these effects take place in the context of the TACRED dataset. Note that the only substitution strategies where this phenomenon needs to be investigated is the \textit{diff.-type} substitution.


\begin{figure*}
  \centering
  \begin{minipage}{\textwidth}
    \fbox{ 
        \parbox{0.96\textwidth}{
            \raggedright
            \small
            \underline{\textbf{Subject sensitive}} \\
            \textbf{residence}: \textit{per:countries\_of\_residence, per:cities\_of\_residence,  per:stateorprovinces\_of\_residence}  \\
            \textbf{headquarter}: \textit{org:country\_of\_headquarters, org:city\_of\_headquarters, org:stateorprovince\_of\_headquarters} \\
            \textbf{death}: \textit{per:city\_of\_death, per:stateorprovince\_of\_death,  per:country\_of\_death} \\
            \textbf{birth}: \textit{per:city\_of\_birth, per:stateorprovince\_of\_birth, per:country\_of\_birth, per:origin} \\ 
            
            \underline{\textbf{Object sensitive}} \\
            \textbf{name}:  \textit{org:alternate\_names,   per:alternate\_names} \\
            \textbf{religion}:  \textit{per:religion,   org:political/religious\_affiliation} \\
            \textbf{member}:  \textit{org:member\_of,   org:top\_members/employees,  per:employee\_of} 
        } 
    }
  \end{minipage}
  \caption{Sets of mutually confusable relations.}\label{fig:mutuallyconfusable}
\end{figure*}

As it turns out, there are certain relations in which the substition of an entity with an entity of another type might generate a different relation in the adversarial sentence compared to the original sentence. We refer to these sets as sets of \textbf{mutually confusable relations}. We list these sets of mutually confusable relations in Fig.~\ref{fig:mutuallyconfusable}. For example, if in a sentence that expresses the \texttt{per:countries\_of\_residence} relation we replace a country with a city, then in the resulting sentence it would make sense that the relation \texttt{per:cities\_of\_residence} is identified.

Figure~\ref{fig:sankey} shows the predictions made for $162$ sentences for the set of \texttt{headquarter} relations for the LUKE model before (on the left) and after the \textit{diff.-type} object entity substitution (on the right). One can observe that the change in prediction to other relations within the same set of mutually confusable relations is not severe. 
It is rarely the case that a sentence where the \texttt{org:country\_of\_headquaters} relation is identified  is transformed into a sentence where the \texttt{org:city\_of\_headquaters} relation is identified~($2$~times), it is rarely the case that a
sentence where 
the \texttt{org:state\-orprovince\-\_of\_headquaters} relation is identified is transformed into a sentence where the \texttt{org:country\_of\_headquaters} relation is identified~($2$~times), and so on.

\begin{figure}
    \centering
    \includegraphics[scale=0.2]{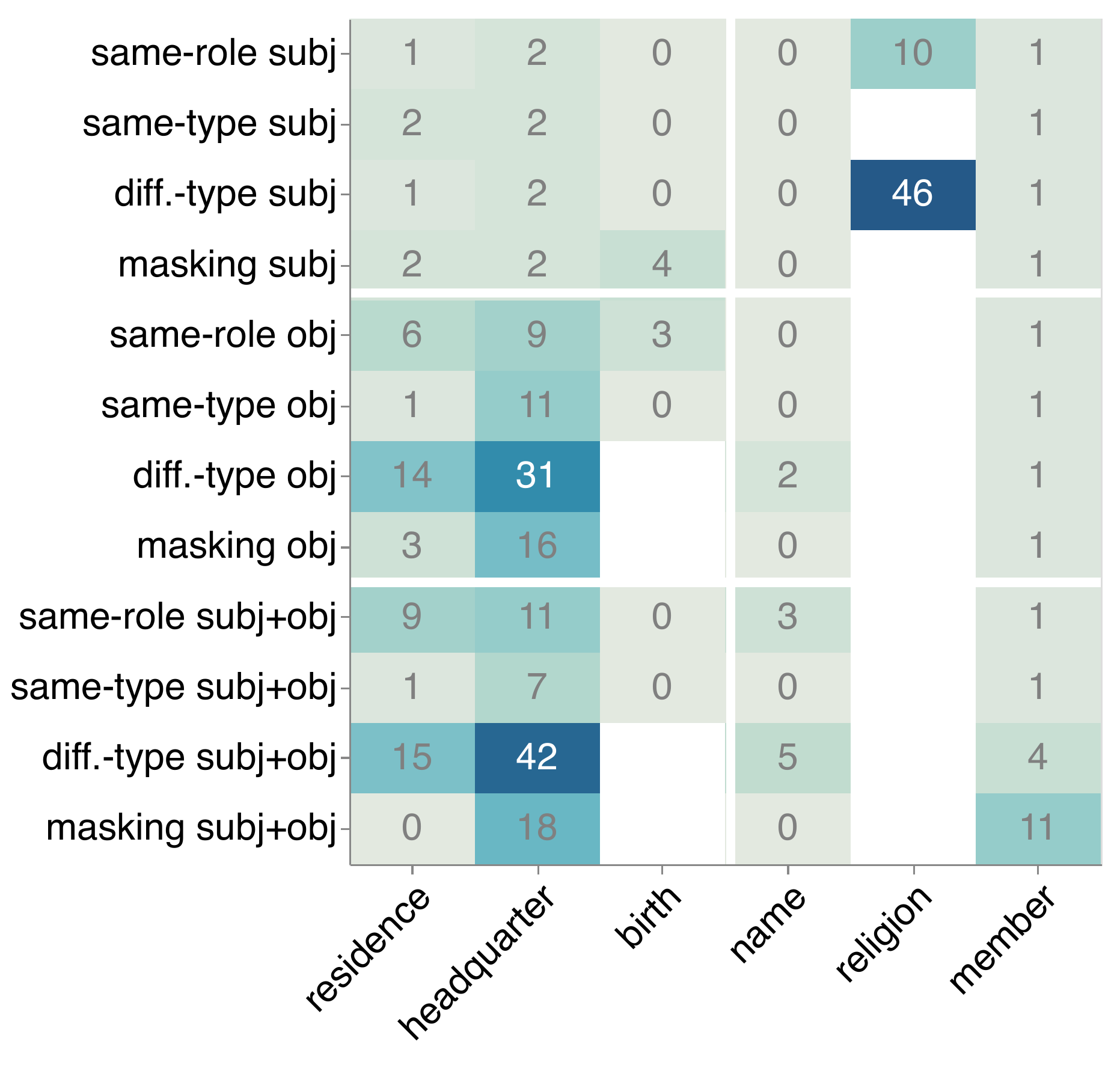}
    \caption{Percentage of predictions within the same set of relations for LUKE.}
    \label{fig:heatmap_in_set_flow}
\end{figure}

We studied this phenomenon for all models and all sets of relations. We computed the percentage of relations that are changed into other relations within the same set. An example for LUKE is shown in Figure~\ref{fig:heatmap_in_set_flow}. The only critical values are for the \textit{diff.-type} subject substitution on the set of \textit{religion} relations and for the \textit{diff.-type} object substitution on the \textit{headquarter} set. However, these values are not representative with $13$ vs. $35$ samples with an average of $98$ samples per cell across the plot. \newline
Thus, although in principle by following the way we evaluate the models one could happen to underestimate the \textit{F1} score of a model due to this phenomenon, we found that this phenomenon rarely occurs for the TACRED datasets, our substitution strategies, and the models we analyzed.

\subsection{Type-aware analysis}\label{subsec:type_aware}
\begin{figure}[!ht]
    \raggedleft
    \caption{Percentages of predicted relations adhering to the type constraints posed by the adversarial examples.}
    \label{fig:perc_constraints}
    \includegraphics[scale=0.34]{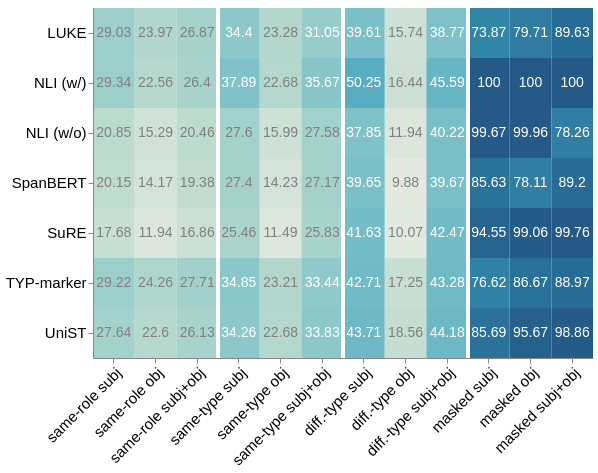}
\end{figure}

Since the work by \citet{Petroni2019LanguageMA}, it is a common assumption that large language models show a strong ability to recall factual knowledge without any explicit access to a knowledge base \cite{Jiang2020HowCW, Liu2021PretrainPA}, e.\thinspace g., by retaining information about entity types and domains/ranges of relations.\newline 
We hypothesize that such an ability can have an impact on the results in adversarial settings. Thus, we investigate whether this sort of information implicitly learned by the models at training time has any impact in the adversarial setting.\newline
While learning this information can be useful for specific tasks (e.\thinspace g., ontology induction), models could find the shortcut to reply to the types of entities and pay less or no attention to the context of the entities in the text. \newline
For instance, consider the following adversarial sample, generated by the \textit{same-role} strategy on subject and object entities:
\begin{center}
\textit{
The first \$ 9 billion in proceeds from the sale will go toward redeeming preferred shares in \underline{Fidelty Investments}\textsubscript{SUBJ:ORGANIZATION}, 
 held by the \underline{NASA}\textsubscript{OBJ:ORGANIZATION} [...]
 }
\end{center}
The correct relation is \texttt{org:parents}, but the combination of entity types \textit{organization} and \textit{organization} also adheres to the domain/range constraints of the following relations: \texttt{org:shareholders}, \texttt{org:alternate\_names}, \texttt{org:member\_of}, \texttt{org:members}, and \texttt{org:subsidiaries}.\newline
If the hypothesis that models inherently learn information about domain and range is accepted, it would follow that, by leveraging such informtion, the models would tend to predict a relation label from this list.\newline
In this case, even if the relation label is incorrect, the models would still show some sort of semantic knowledge on entity tpes.\newline
To test this hypothesis, we first create a list of all the possible subject type and object type combinations that occur in the training split, and the relation they are involved in.\newline
Then, for every example in the adversarial datasets, we check whether the predicted relation falls into one of the possible labels according to the domain and range of the generated sentence.\newline
If a combination of entity types is never found in the training set (for instance when at least one entity is masked), the only available relation is \texttt{no\_relation}.\newline
The percentages of predicted relations that adhere to the constraints posed by the newly generated sentences are shown in Figure~\ref{fig:perc_constraints}.
According to the results, the initial hypothesis that the investigated models use information about entity types is proved wrong. For instance, for the example shown above, all the models, except for \texttt{TYP\_marker}, predicted the relation shown in the sentence as being \texttt{no\_relation}.\newline
As shown in Figure~\ref{fig:perc_constraints}, this is not an anomaly. In fact, models only adhere to constraints posed by domain/range by $50\%$ at most, but in most cases much lower, with the only exception being in the case of \textit{masking} adversarials. However, this is expected since in this case the only available relation is \textit{no\_relation}.\newline
For the other strategies of substitution, the models weakly adhere to entity types combination, even when entity types are not changed by the adversarial strategy (i.\thinspace e., for \textit{same-role} and \textit{same-type} adversarials).\newline
Similarly, since the substitutions are randomly generated, it might happen for the \textit{diff.-type} strategy that the generated adversarials adhere to a different relation than the one holding in the original sentence (see Section~\ref{sec:limitations}). Even in those cases, the models adhere to the newly posed domain/range constraints with a low frequency.\newline
At best, NLI (w/) adheres to such constraints 50.25\% of the time, for \textit{diff.-type} substitution on subject entities, with all the other models faring worse than this.\newline
We can thus assume that models take into account entities' surface forms rather than implicit information about entity types and domain/range.\newline
As mentioned already in Section~\ref{subsec:general_results}, in this case as well, the models are less likely to adhere to known entity types constraints when the object entity is the one being substituted. This is clearly seen in the \textit{diff.-type} substitution strategy, where the examples in which the objects are substituted are, on average, $66.16\%$ less likely to adhere to constraints posed by entity types.


\section{Conclusion and Future Work}\label{sec:conclusion}In this work, we test a set of SOTA Relation Extraction models on multiple adversarial datasets created through semantically motivated substitution strategies. As shown by the results, the models are heavily affected by this kind of adversarials, and in particular, they show an overreliance on the surface form of entities to predict the correct relation label, rather than relying on the actual linguistic structure of the relation expressed in a sentence or on the entities' types.\newline
The models lose an average of $48.5\%$ of \textit{F1} score when facing adversarials, with certain adversarial strategies being more impactful than others.\newline
The possibility that models are led astray by entity types is proven false by a deeper analysis of the results. In fact, models generally default to the \texttt{no\_relation} class even in the face of adversarials where a relation between the two entities might exist, according to their types. While, given the unbalanced nature of the TACRED dataset, this is to be expected, it also means that models are easily led astray by changes in the input data, even in cases that leave the relation unchanged.\newline
In future work, applying the adversarial methodology to another dataset might prove useful in testing whether models trained on another set of data might be more resilient to adversarial attacks.\newline
Furthermore, while semantically motivated adversarials have proven to be effective when evaluating RE models, it is possible to build \textit{syntactically} motivated adversarials that substitute or mask specific parts-of-speech in sentences, e.\thinspace g., verbs, to test models' reliance on such features.\newline
Such adversarials can be included in future work, to prove whether models actually tend to rely on entity's surface forms, or are just led astray by any kind of unexpected change in the testing set.\newline
While this work focuses on the use of adversarials as a means to explore the workings of language models, adversarials have also been proposed as a way to train more robust models \cite{szegedy:14}. Despite being out of scope for the present paper, an investigation of how model performance improves / robustness increases when trained on the generated adversarials can help to gain a deeper understanding of the models.

\section{Limitations}\label{sec:limitations}
We outlined that the adversarial strategy \textit{diff.-type} might generate samples expressing another relation where two relations are expressed by similar structures. An example and a case study analyzing the models' behavior is shown in Section~\ref{subsec:rel}.

\begin{figure}[!ht]
    \centering
    \includegraphics[scale=0.45,trim=7 7 7 7, clip]{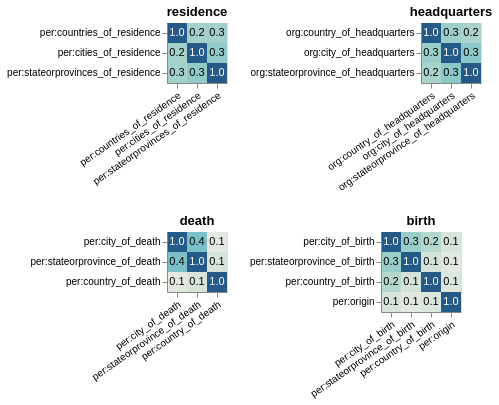}
    \caption{Jaccard similarity between groups of similar relations}
    \label{fig:heatmap_jaccard}
\end{figure}
To investigate the amount of such possible false negatives independently of the substitution strategy for TACRED, we compute the Jaccard similarity on the token basis for sentences within groups of mutually confusable relations. To compute this similarity, entity mentions were removed from every sentence. The results are shown in Figure~\ref{fig:heatmap_jaccard}.\newline
Since sentences referring to different relations have, in general, an average of one third of their tokens in common~(this also takes into account function words), it can be assumed that the actual overlap between sentences expressing different relations is low enough for false negatives not to be considered an issue.\newline
We further checked how many times the \textit{diff.-type} strategy generates sentences representing similar relations. To do so, we compute how many sentences generated from one relation generate sentences representing another, similar one. These percentages are shown in Table \ref{tab:perc_critical}.

\begin{table}[!ht]
    \centering
    \caption{Percentages of generated sentences that represent similar relations}
    \label{tab:perc_critical}
    \footnotesize
    \begin{tabular}{l|l}
        \textbf{Relation} & \textbf{\%} \\
        \hline
         \texttt{per:cities\_of\_residence} & 4.0 \\
         \texttt{per:countries\_of\_residence} & 4.51 \\
         \texttt{per:stateorprovinces\_of\_residence} & 2.38 \\
         \texttt{per:origin} & 3.47 \\
         \texttt{per:cities\_of\_death} & 4.17 \\
         \texttt{per:country\_of\_death} & 4.17 \\
         \texttt{per:stateorprovince\_of\_death} & 1.19 \\
         \texttt{per:stateorprovince\_of\_birth} & 2.78 \\
         \texttt{per:country\_of\_birth} & 6.67 \\
         \texttt{per:city\_of\_birth} & 1.67 \\
         \texttt{org:stateorprovince\_of\_headquarters} & 5.36 \\
         \texttt{org:country\_of\_headquarters} & 5.66 \\
         \texttt{org:city\_of\_headquarters} & 3.65 
    \end{tabular}

\end{table}

Even though the substitutions might generate a correct relation in some cases, we have already shown that the models tend not to agree to domain/range information in the newly generated sentences, and this is the case for all models across all types of generated adversarials.\newline

\section{Ethics}
Our work primarily focuses on the investigation of existing language models, which in itself does not introduce new ethical biases. Instead, our work contributes to the broader conversation about the ethical development of language models, emphasizing the need for transparency and fairness.\newline
Nevertheless, when considering the use of adversarials for training, it is crucial to ensure that such manipulations do not introduce new harmful biases. By default, our methodology might generate false facts, which, if used to train a model, would then be learned by the model. \newline
Finally, since the TACRED dataset is based on the English language, this investigation might show different results with different languages, and as such should not be taken for granted in a multilingual setting.


\section{References}
\bibliographystyle{lrec-coling2024-natbib}
\bibliography{custom}

\clearpage


\end{document}